\documentclass[fleqn,10pt]{wlscirep}
\usepackage[utf8]{inputenc}
\usepackage[T1]{fontenc}
\usepackage{hyperref}
\hypersetup{colorlinks,%
citecolor=black,%
filecolor=black,%
linkcolor=black,%
urlcolor=black
}
\usepackage{memhfixc}
\usepackage{lineno}

\title{FlowRoI: A Fast Optical-Flow–Driven Region-of-Interest Extraction Framework for High-Throughput Image Compression in Immune Cell Migration Analysis}

\author[1,2]{Xiaowei Xu}
\author[1]{Justin Sonneck}
\author[3]{Hongxiao Wang}
\author[4]{Roman Burkard}
\author[4,5]{Hendrik Wöhrle}
\author[4]{Anton Grabmasier}
\author[1,6*]{Matthias Gunzer}
\author[1,*]{Jianxu Chen}

\affil[1]{Leibniz-Institut für Analytische Wissenschaften– ISAS– e.V., Dortmund, Germany}
\affil[2]{Guangdong Provincial People's Hospital, Guangdong Academy of Medical Sciences, Guangzhou, China, 510080}
\affil[3]{Academy for Multidisciplinary Studies, Capital Normal University, Beijing, China}
\affil[4]{Department of Electronic Components and Circuits, University of Duisburg-Essen, Duisburg, Germany}
\affil[5]{Fraunhofer Institute for Microelectronic Circuits and Systems, Duisburg, Germany}
\affil[6]{Institute for Experimental Immunology and Imaging, University Hospital, University of Duisburg-Essen, Essen, Germany}

\affil[*]{Corresponding authors}


\begin{abstract}
Autonomous migration is essential for the function of immune cells such as neutrophils and plays a pivotal role in diverse diseases. Recently, we introduced ComplexEye, a multi-lens array microscope comprising 16 independent aberration-corrected glass lenses arranged at the pitch of a 96-well plate, capable of capturing high-resolution movies of migrating cells. This architecture enables high-throughput live-cell video microscopy for migration analysis, supporting routine quantification of autonomous motility with strong potential for clinical translation.
However, ComplexEye and similar high-throughput imaging platforms generate data at an exponential rate, imposing substantial burdens on storage and transmission. To address this challenge, we present FlowRoI, a fast optical-flow–based region-of-interest (RoI) extraction framework designed for high-throughput image compression in immune cell migration studies. FlowRoI estimates optical flow between consecutive frames and derives RoI masks that reliably cover nearly all migrating cells. The raw image and its corresponding RoI mask are then jointly encoded using JPEG2000 to enable RoI-aware compression.
FlowRoI operates with high computational efficiency, achieving runtimes comparable to standard JPEG2000 and reaching an average throughput of about 30 frames per second on a modern laptop equipped with an Intel i7-1255U CPU. In terms of image quality, FlowRoI yields higher peak signal-to-noise ratio (PSNR) in cellular regions and achieves 2.0–2.2× higher compression rates at matched PSNR compared to standard JPEG2000.
To further assess downstream utility, we evaluated FlowRoI using cell instance segmentation as a representative task. At comparable segmentation accuracy (average precision), FlowRoI enables 2.2–2.3× higher compression rates. Additionally, FlowRoI is training-free, requires only a small number of hyperparameters, and demonstrates robust performance across parameter settings. Owing to its high speed, appropriate parameters can be easily identified through a lightweight search in practice.

\end{abstract}
\begin{document}

\flushbottom
\maketitle
%
%
\thispagestyle{empty}


\section*{Introduction}
Autonomous migration is essential for the function of immune cells and plays a central role in numerous diseases \cite{roos1993novel,mckinney2020metabolic,bornemann2020defective,kuhns2016cytoskeletal}.
Among these cells, neutrophils—the most abundant leukocytes in human blood—act as critical first responders \cite{filippi2019neutrophil}.
They rapidly migrate into inflamed tissues within minutes, enabling immediate host defense \cite{peters2008vivo}.
However, this same migratory capacity can be detrimental: neutrophil infiltration not only protects against infection but may also aggravate pathology.
For example, their recruitment into tumors correlates with poor prognosis \cite{gentles2015prognostic}, and excessive migration into ischemic heart \cite{merz2019contemporaneous} or brain \cite{neumann2015very} tissue exacerbates sterile injury.
Thus, autonomous migration represents a double-edged sword, capable of conferring both protection and harm.

Decades of work have uncovered molecular mechanisms that govern neutrophil migration \cite{kienle2021neutrophils}.
The bacterial peptide fMLP, for instance, potently induces chemotaxis toward infection \cite{metzemaekers2020neutrophil}, yet structurally similar ligands originating from host mitochondria can redirect neutrophils into sterile inflammatory sites.
Depending on context, such migration may support tissue repair—e.g., by promoting revascularization—or worsen pathology, as observed in stroke and cancer.
Consequently, distinguishing and selectively modulating these divergent migratory programs is of considerable therapeutic interest.
However, given the diversity of chemotactic ligands, predicting modulatory effects computationally is nearly impossible, necessitating systematic high-throughput experimental approaches.
High-throughput migration assays enable parallel, large-scale functional profiling, facilitating the discovery of selective modulators that might, for example, prevent neutrophil entry into tumors while preserving antimicrobial defense.

To address this need, we recently introduced ComplexEye \cite{cibir2023complexeye}, a multi-lens video microscope that integrates 16 aberration-corrected glass lenses, each equipped with an individual detector and illumination path.
This configuration allows simultaneous imaging of 16 wells of a 96-well plate, or 64 wells of a 384-well plate, at one frame every 8 seconds.
ComplexEye enables energy-efficient, high-throughput migration analysis across hundreds of conditions.
We demonstrated that the system can process multiple clinical samples in parallel and can screen 1,000 compounds to identify 17 modulators of human neutrophil migration within four days—a task that would require approximately 60 times longer using a conventional video microscope.

However, high-throughput immune cell migration imaging generates data at an unprecedented scale, intensifying the long-standing storage challenge in microscopy.
Over the past decade, the widespread adoption of digital imaging and advances in multidimensional microscopy have driven explosive data growth, with many core facilities now producing petabytes of data annually \cite{giepmans2023focus,way2023evolution,poger2023big}.
A single high-content screening experiment may generate hundreds of thousands of images \cite{massei2025high}.
In our ComplexEye platform, a one-hour experiment produces 28,800 images (about 16 GB).
In industrial-scale settings, yearly data volumes easily reach the petabyte range \cite{way2023evolution}.
The storage and transmission of such massive datasets are both time-consuming and costly, underscoring the urgent need for efficient image compression tailored to high-throughput migration analysis.

Here, we introduce FlowRoI, a fast optical-flow–based region of interest (RoI) extraction framework designed for high-throughput image compression in immune cell migration studies.
FlowRoI first computes optical flow between adjacent frames and derives RoI masks that capture nearly all migrating cells.
The raw image and the corresponding RoI mask are then jointly encoded using JPEG2000 to enable RoI-based compression.
FlowRoI operates extremely efficiently, achieving about 30 frames per second on a standard laptop with an Intel i7-1255U CPU.
In cell-containing regions, FlowRoI yields higher peak signal-to-noise ratio (PSNR), and at matched PSNR it achieves 2–2.2× higher compression rates compared to standard JPEG2000.

To validate its downstream utility, we used cell instance segmentation as a representative analysis task.
At comparable segmentation performance, FlowRoI achieves 2.2–2.3× higher compression rates.
Importantly, FlowRoI is training-free, requires only a small set of hyperparameters, and remains robust across a wide range of parameter choices.
Even when a small number of cells are not captured in the RoI masks, we show that downstream performance remains unaffected, as these cells are typically too challenging for state-of-the-art methods to detect in the first place.
Finally, we provide an analysis of key hyperparameters, offering guidance for practical deployment in high-throughput imaging workflows.


\section*{ComplexEye platform}
ComplexEye \cite{cibir2023complexeye} is a multi-lens, high-throughput video microscope designed for embedded and parallel analysis of immune cell migration.
The system architecture comprises five major components: an illumination module, a standard multi-well plate, a lens array, a field-programmable gate array (FPGA) board, and a high-precision motorized stage.
The illumination module uses a single light source delivered through fiber optics, with each well receiving homogeneous illumination via an individual Köhler optic, forming a Köhler optic array.
This design ensures consistent, high-quality lighting across plates of up to 384 wells.
The Köhler optic array, lens array, and FPGA board are mechanically coupled, enabling them to translate as a single rigid unit relative to the well plate.
The FPGA board performs two primary functions: image acquisition—by receiving data streams from CMOS sensors and forwarding them for downstream processing and storage—and motor control, where it regulates the stage system to drive coordinated movement of the coupled assembly.
During operation, the FPGA-controlled assembly moves periodically, allowing ComplexEye to capture one image per well every 8 seconds.
Consequently, each well generates a time-resolved sequence suitable for high-throughput migration analysis in both 96- and 384-well formats.
Additional details on the optical design, electronic architecture, and system performance can be found in the original ComplexEye publication \cite{cibir2023complexeye}.

\begin{figure*}[htb]
\includegraphics[width=\textwidth]{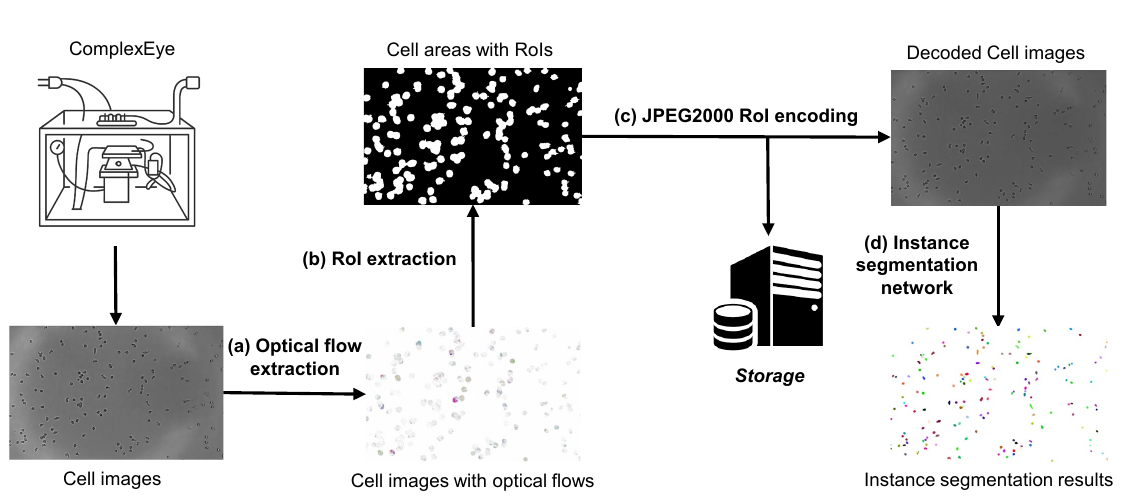}
\caption{Framework of FlowRoI for high-throughput image compression in immune cell migration analysis.} 
\label{framework}
\end{figure*}

\section*{Method}
For downstream migration analysis, precise localization of cells within each frame is essential.
However, most pixels in an immune cell image correspond to background regions that do not contribute meaningful biological information.
The core idea of FlowRoI is to identify and isolate these task-relevant cellular regions and allocate higher encoding priority to them during compression.
The overall framework of FlowRoI is illustrated in Figure \ref{framework} and consists of four main stages.
In the first stage, optical flow is computed from the ComplexEye image sequence using pairs of adjacent frames.
This captures the local motion patterns associated with migrating cells.
In the second stage, regions of interest (RoIs) are derived from the resulting optical-flow field.
Here, RoIs correspond to cell-containing regions, which are critical for downstream tasks such as semantic segmentation, instance segmentation, cell tracking, and motion quantification.
In the third stage, JPEG2000 RoI encoding is applied.
RoI pixels are encoded with higher priority—meaning lower compression—while background pixels receive lower priority.
This differential treatment allows the compressed bitstream to retain high fidelity in biologically meaningful regions while substantially reducing overall data size.
The compressed output can be stored or transmitted efficiently.
In the final stage, the encoded images are decoded and subsequently used for downstream analysis.
To assess the effectiveness of FlowRoI, we employ cell instance segmentation as a representative biological task.

\subsection*{Optical flow extraction}
Optical-flow extraction consists of three sequential refinement stages: denoising, stabilization, and flow estimation.
Image noise is reduced using a two-step denoising strategy that combines median filtering and bilateral filtering to suppress high-frequency noise while preserving cellular boundaries.
To correct global translational drift introduced during image acquisition, each frame is then registered to its predecessor via phase-correlation–based alignment.
After denoising and alignment, pixel-wise motion is estimated using the dense Farnebäck optical-flow algorithm \cite{farneback2001very}, which produces displacement vectors describing local cellular and subcellular motion.
This pipeline ensures that optical flow is computed from noise-reduced and temporally aligned image pairs, yielding robust and accurate flow fields for subsequent RoI extraction.

\subsection*{Region of Interest extraction}
RoI extraction from the flow field follows a saliency-based segmentation procedure.
A motion-saliency map is first constructed by combining flow magnitude and spatial gradients with a weighting factor of 0.7. This enhances regions exhibiting coherent local motion while suppressing isolated noise responses. 
The saliency map is then normalized to highlight dynamically active regions across the sequence.
Pixels exceeding a quantile-based threshold (the RoI threshold) are selected to generate an initial binary mask.
Morphological opening and closing are applied to remove small spurious components and to fill small gaps. Further refinement is performed by removing residual small connected components based on area criteria to ensure that only sufficiently large, biologically meaningful regions are retained.
When motion between adjacent frames is weak, FlowRoI optionally incorporates RoI masks from neighboring frames, producing a more stable and complete mask.
The overall process yields a clean and compact binary RoI mask that delineates motion-sensitive regions corresponding to the majority of cells.

\begin{table*}[tb]
\centering
\caption{Hyperparameters in FlowRoI and their illustrations.}
\begin{tabular}{@{}lll@{}}
\toprule
{Parameter name} & Illustration & Values \\ \midrule
Denoise & Whether denoise is performed in optical flow extraction & On or off \\
RoI threshold & The rate of pixels in optical flow remained for further analysis used in RoI extraction & (0, 1) \\
Adjacent factor & The number of adjacent masks considered for the ensemble used in RoI extraction & 0,1,2 or more \\
Scaling factor& the relative importance of the RoI over the background in JPEG2000 RoI encoding & [1, 10]\\
Compression rate& The target compression rate used in JPEG2000 RoI encoding & numbers larger than 1
\\ \bottomrule
\end{tabular}
\label{table_parameters}
\end{table*}

\subsection*{Implementation details}
Following optical-flow estimation and RoI extraction, the original image and corresponding RoI mask are encoded using JPEG2000 \cite{kadudu}.
JPEG2000 RoI compression allows different compression strengths to be applied to different image regions, preserving higher fidelity in important areas while retaining strong overall compression.
Among the available RoI tools in the JPEG2000 standard, FlowRoI employs the scaling-based RoI method, in which a user-controlled scaling factor specifies the relative importance of RoI pixels.
For downstream evaluation, decoded images are processed using a modern instance segmentation network.

FlowRoI includes five hyperparameters, summarized in Table \ref{table_parameters}.
Denoising is optional and may be enabled or disabled depending on the imaging setup. Although denoising may remove a small number of very low-contrast cells, our experiments indicate that these cells occur infrequently and are typically too challenging even for state-of-the-art deep networks to detect reliably. Users may adjust this parameter based on their specific data characteristics.
The RoI threshold controls the proportion of high-saliency pixels retained as RoI. For example, a threshold of 0.2 selects the top 20\% of saliency values.
The adjacent factor determines whether RoI masks from neighboring frames are incorporated to stabilize the current mask.
The remaining hyperparameters—the scaling factor and compression rate—correspond to JPEG2000 RoI settings, where a larger scaling factor increases RoI priority during encoding and a larger compression rate results in stronger compression and smaller file size.

\begin{figure}
\centering
\includegraphics[width=0.99\textwidth]{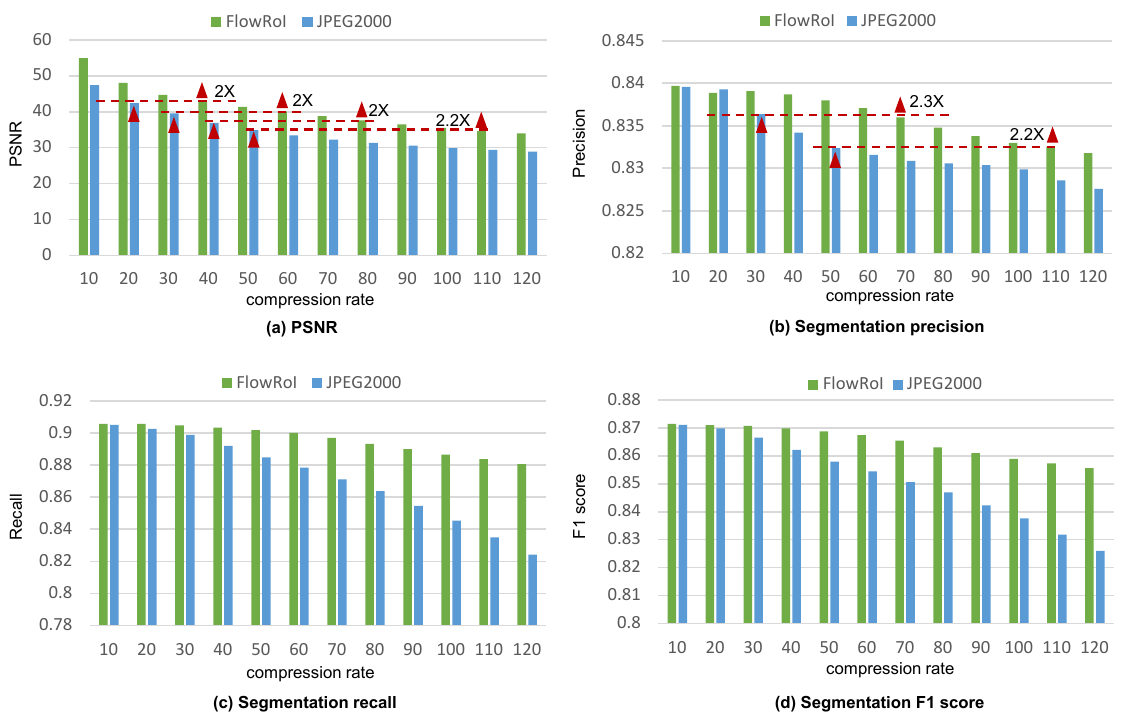}
\caption{Comparison of FlowRoI and JPEG2000 in image quality using PSNR and instance segmentation using precision, recall and F1 score.
} 
\label{fig_cr}
\end{figure}

\begin{figure}
\centering
\includegraphics[width=0.8\textwidth]{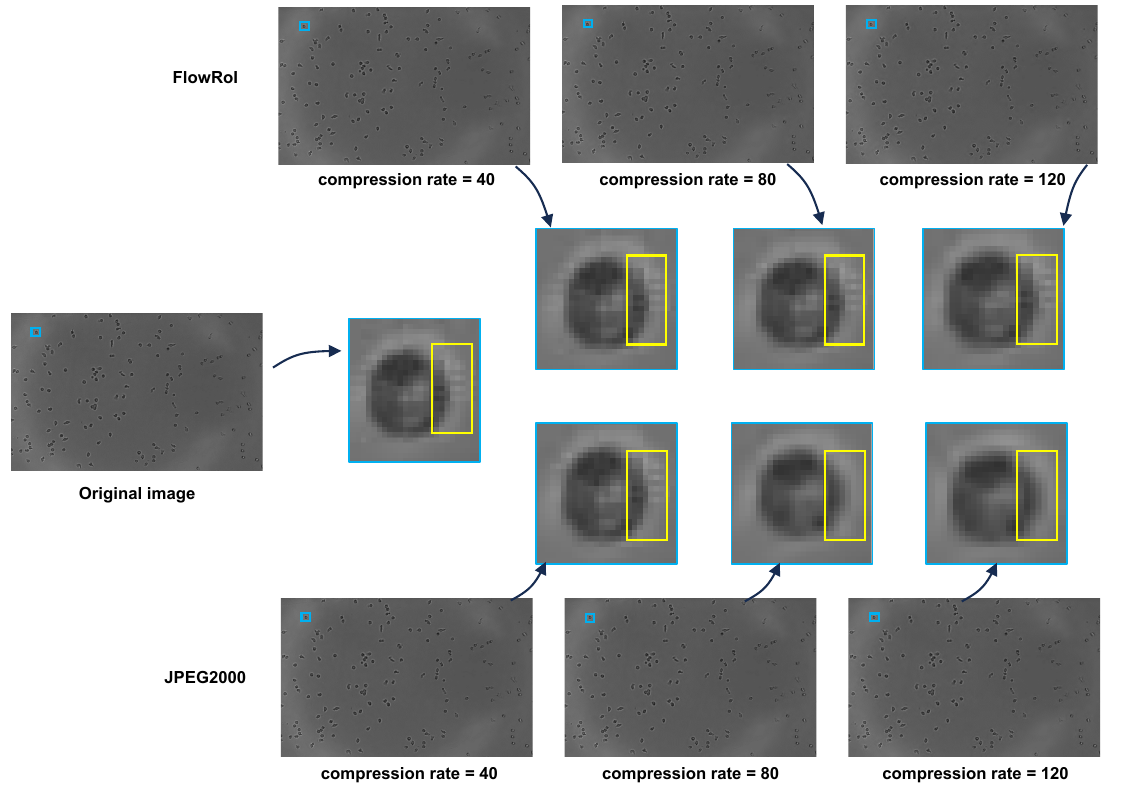}
\caption{Qualitative comparison of FlowRoI and JPEG2000 with different compression rate.} 
\label{fig_qualitative_results}
\end{figure}

\begin{figure}[htb]
\centering
\includegraphics[width=0.96\textwidth]{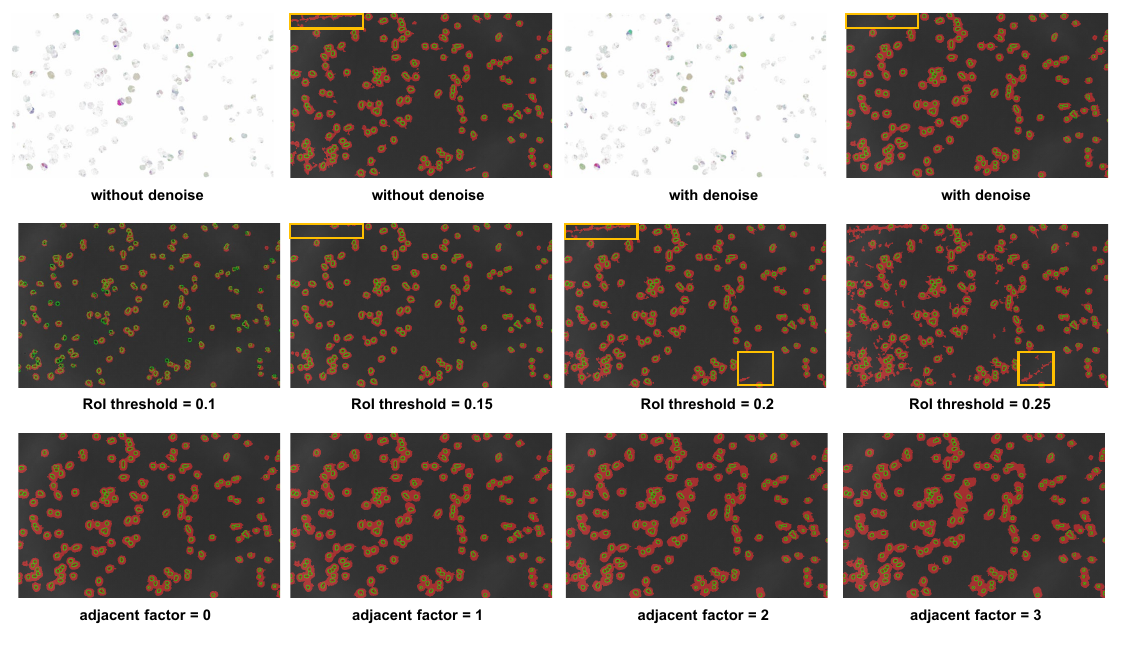}
\caption{Qualitative intermediate outputs with different settings of denoise, RoI threshold, and adjacent factors. After applying denoising, several scattered and irregular artifacts (highlighted in yellow) are effectively removed, resulting in cleaner RoI masks without affecting the main cell regions. } 
\label{fig_process_results}
\end{figure}

\begin{figure}[!h]
\centering
\includegraphics[width=\textwidth]{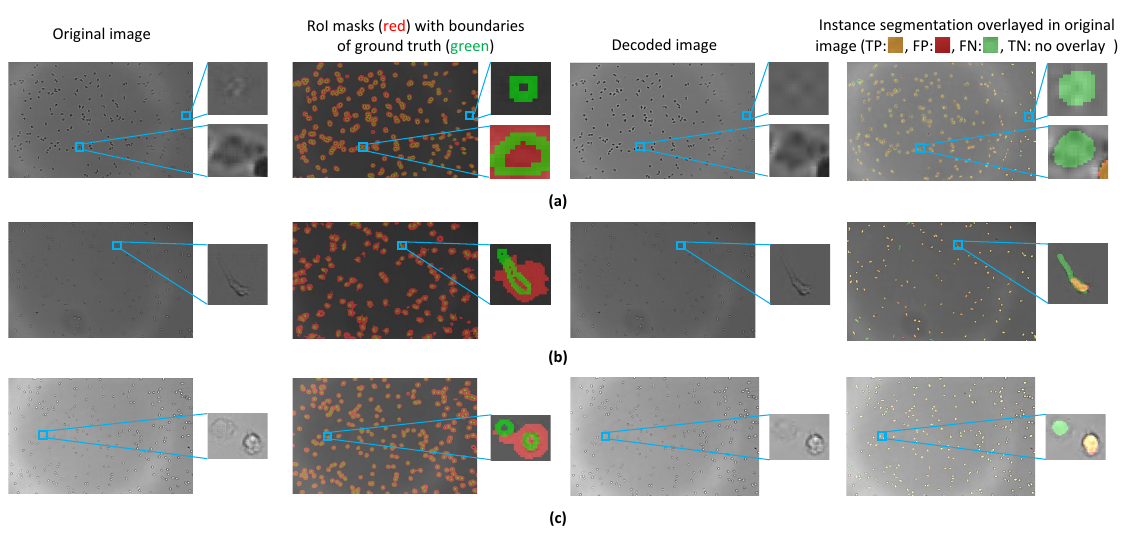}
\caption{Examples of cell missing cases (TP: true positive, FP: false positive, FN: false negative, and TN: true negative).} 
\label{fig_RoI_missing_cases}
\end{figure}
 
\section*{Results and discussion}

\subsection*{Experimental setup}
We evaluated FlowRoI on a dataset containing six ComplexEye videos with a total of 2,628 2D images \cite{dataset_cell_track}.
All images were manually annotated by experts for instance segmentation. FlowRoI was implemented and tested on a laptop equipped with an Intel i7-1255U CPU.
For instance segmentation, we adopted Cellpose-SAM \cite{pachitariu2025cellpose} and fine-tuned it using 240 annotated images.
The backbone sam model was selected, with a batch size of 2, learning rate of 0.02, and 400 training epochs. Training was performed using the PyTorch framework \cite{paszke2019pytorch} on a single NVIDIA RTX A100 GPU.
During inference, all configurations used identical model parameters.
For comparison, we selected JPEG2000 \cite{taubman2002jpeg2000} as a representative state-of-the-art compression method.
For instance segmentation evaluation, we report precision, recall, and F1-score at an IoU threshold of 0.5.
It is important to note that we use precision(IoU = 0.5) rather than average precision, as our evaluation is performed at a single IoU threshold without sweeping across confidence levels.
All three metrics are derived from the numbers of true positives (TP), false positives (FP), and false negatives (FN), computed using the widely adopted IoU threshold of 0.5 \cite{pachitariu2025cellpose}.

\begin{figure}[htb]
\centering
\includegraphics[width=\textwidth]{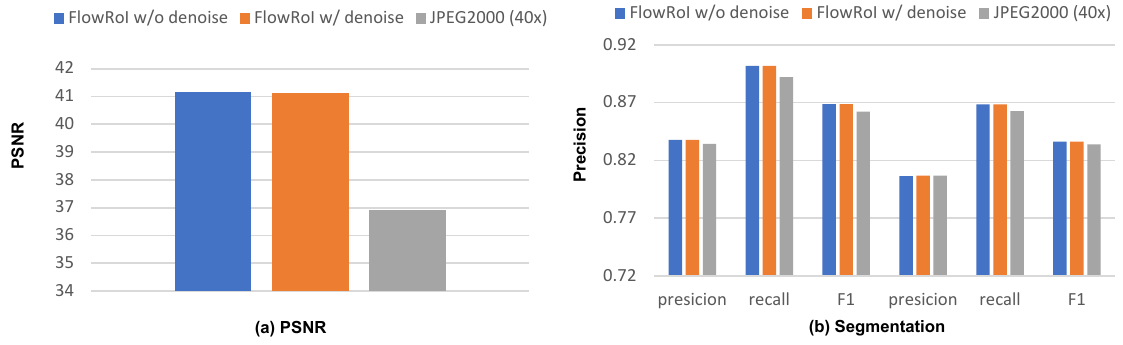}
\caption{Parameter discussion of denoise. RoI threshold: 0.2, adjacent factor: 0, scaling factor: 5, compression rate: 40, and JPEG2000 (40x): JPEG2000 compression with a compression rate of 40.} 
\label{fig_denoise}
\end{figure}

\begin{figure}[htb]
\centering
\includegraphics[width=\textwidth]{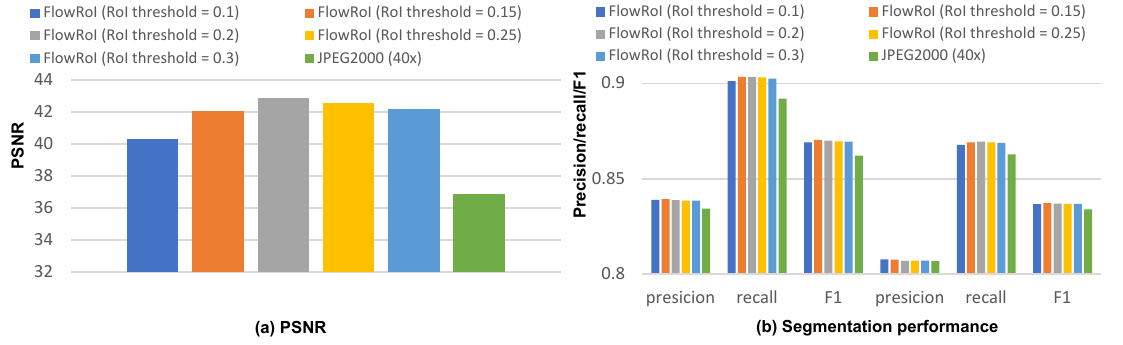}
\caption{Parameter discussion of RoI threshold. Denoise: 1, Adjacent factor: 0, scaling factor: 5, compression rate: 40, and JPEG2000 (40x): JPEG2000 compression with a compression rate of 40.} 
\label{fig_threshold}
\end{figure}

\begin{figure}[!htb]
\centering
\includegraphics[width=0.99\textwidth]{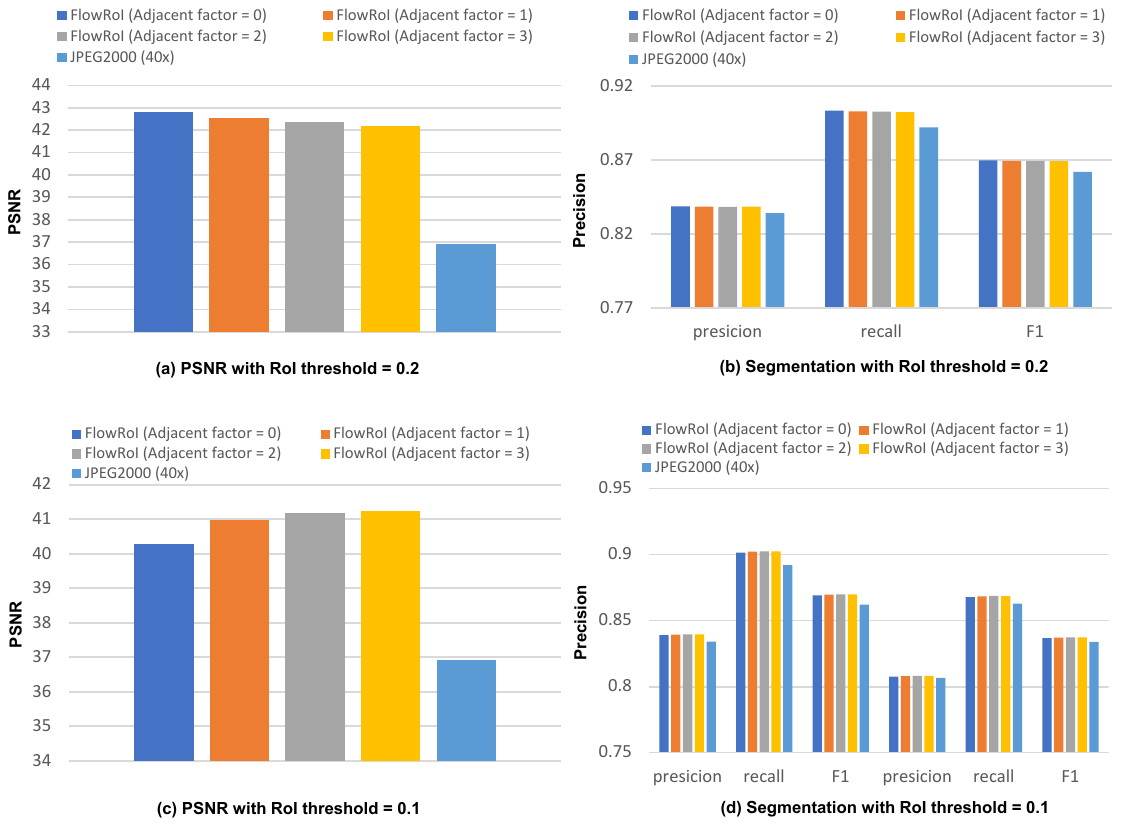}
\caption{Parameter discussion of adjacent factor. Denoise: 1, RoI threshold: 0.2, scaling factor: 5, compression rate: 40, and JPEG2000 (40x): JPEG2000 compression with a compression rate of 40.} 
\label{fig_merge_step}
\end{figure}

\begin{figure}[htb]
\centering
\includegraphics[width=\textwidth]{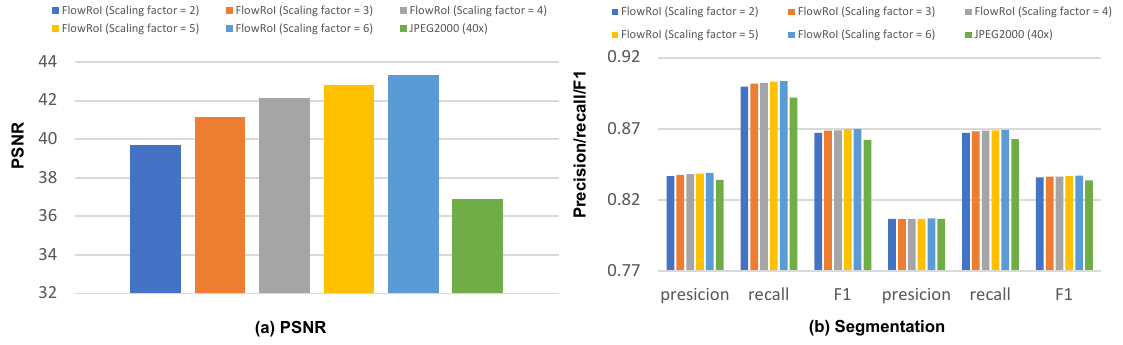}
\caption{Parameter discussion of scaling factor. Denoise: 1, RoI threshold: 0.2, adjacent factor: 0, compression rate: 40, and JPEG2000 (40x): JPEG2000 compression with a compression rate of 40.} 
\label{fig_importance}
\end{figure}

\subsection*{Overall performance}
Figure \ref{fig_cr} compares FlowRoI and JPEG2000 in terms of image quality and downstream instance segmentation across a range of compression rates.
Both methods show gradual declines in PSNR and segmentation performance as compression increases, with sharper degradation at higher compression levels.
Importantly, the performance drop occurs later for FlowRoI than for JPEG2000, indicating better robustness of FlowRoI in preserving task-relevant information under strong compression.
The fine-tuned Cellpose-SAM model achieves a precision above 0.80—consistent with the results reported in the original publication \cite{pachitariu2025cellpose}—confirming that our retraining procedure was correctly performed.
Across all compression rates (Fig. 3a), FlowRoI consistently achieves higher PSNR than JPEG2000.
At equal PSNR, FlowRoI provides a 2–2.2× improvement in compression efficiency.
Similarly, for instance segmentation precision (Fig. 3b), FlowRoI yields 2.2–2.3× higher compression efficiency at comparable accuracy levels.

A qualitative comparison is shown in Figure \ref{fig_qualitative_results}. At the full-image scale, both methods appear visually similar.
However, after zooming into local regions (highlighted in yellow), FlowRoI preserves fine cellular texture and boundary morphology across all compression rates, even at a compression rate of 120.
JPEG2000 preserves similar detail only at a compression rate of 40; at higher rates (80 and 120), fine structures are blurred or lost.

Figure 5 illustrates intermediate outputs under varying denoising, RoI thresholds, and adjacent factors.
Denoising removes scattered artifacts without altering major cell regions.
The RoI threshold controls how many pixels are retained: higher thresholds enlarge RoI masks from cell centers outward, but excessively high values may introduce background noise.
The adjacent factor governs temporal merging across frames: larger values expand RoI masks by integrating temporal footprints of moving cells.
Because FlowRoI runs extremely efficiently (30 FPS on an Intel i7-1255U CPU), users can easily sweep parameter combinations to find an optimal balance between RoI coverage and noise suppression.

\subsection*{Cell missing cases}
Although FlowRoI aims to cover all cells while keeping RoI size minimal, some cells may still be missed (Fig. \ref{fig_RoI_missing_cases}).
These missed cells typically exhibit very low contrast and appear similar to the background, resulting in weak motion flow and exclusion from RoI masks.
During compression, such regions are encoded with the same priority as the background, further reducing their visibility after decoding.
Consequently, these cells may not be recognized during instance segmentation.
Notably, even in the original uncompressed images, such low-contrast cells are not reliably detected by Cellpose-SAM.
For example, as shown in the first row of Figure \ref{fig_RoI_missing_cases}, the leftmost missed cell has higher contrast than the others but still fails to be segmented correctly.
Overall, the number of such cells is extremely small (107 out of 409,968 total cells) and has a negligible impact on quantitative performance.

\subsection*{Parameter discussion}

Figure \ref{fig_denoise} shows the effect of denoising.
FlowRoI with and without denoising performs similarly, as the removed noise components are minor.
Both versions significantly outperform JPEG2000.
Although the improvement is small, denoising is generally recommended because it stabilizes the results.

Figure \ref{fig_threshold} illustrates performance at different RoI thresholds.
PSNR exhibits an inverse-U trend, peaking at a threshold of 0.2.
Instance-segmentation performance shows a different pattern: it increases from thresholds of 0.1 to 0.2, but remains nearly constant for larger values.
A likely explanation is that at low thresholds (e.g., 0.1), some cells are not fully included in the RoI (Fig. \ref{fig_process_results}), reducing both PSNR and segmentation accuracy.
As the threshold increases, more cells are covered—even if only partially—improving both metrics.
When thresholds become large, RoI masks grow too large, leaving fewer bits to encode each RoI pixel and decreasing PSNR.
However, segmentation performance remains unaffected because slight blurring in <5\% of the image does not impact IoU-0.5 segmentation.

Figure \ref{fig_merge_step} shows the influence of the adjacent factor.
With a high RoI threshold (0.2), increasing the adjacent factor reduces PSNR and segmentation performance by adding excess background to RoI masks.
With a low RoI threshold (0.1), increasing the adjacent factor improves both PSNR and segmentation performance by compensating for missed cells via temporal integration.

The impact of the scaling factor is shown in Figure \ref{fig_importance}.
Higher scaling factors increase both PSNR and segmentation performance, as expected.
However, PSNR increases more than segmentation accuracy, because once image quality surpasses the threshold required for robust segmentation, further enhancement yields diminishing returns.

\begin{table}[!tb]
\centering
\caption{Discussion of runtime and performance with related methods.}
\begin{tabular}{@{}llll@{}}
\toprule
Method & Training & Platform & PSNR gain over JPEG2000 \\ \midrule
RoI compression\cite{han2006image} & Yes & $\textit{CPU}$ & around 3 dB\\
Swin transformer based RoI compression\cite{li2023RoI} & Yes & GPU & 2-2.5 dB\\
RoI compression with variable-rate compression\cite{kao2023transformer} & Yes & GPU & around 3-4 dB\\
Variable-rate deep image compression\cite{song2021variable} & Yes & CPU & around 2 dB\\
Image codec paradigm\cite{chen2021new} & Yes& GPU& around 5 dB\\
Content-weighted image compression\cite{li2018learning} & Yes& GPU& no significant improvement \\
Saliency segmentation based compression\cite{li2024saliency} & Yes& GPU & 10\% bitrate reduction \\
FlowRoI & $\textit{NO}$& $\textit{CPU}$ & around 5 dB (50-55\% birrate reduction) \\
\bottomrule
\end{tabular}
\label{table_comparison}
\end{table}

\subsection*{Discussion with related works}
RoI-based image compression has attracted considerable interest over the past decades.
Table \ref{table_comparison} summarizes comparisons with related methods.
FlowRoI requires no training, runs quickly (0.14 s per case on an Intel i7-1255U CPU using a single thread; JPEG2000 requires 0.09 s under the same conditions), and offers substantial performance gains over JPEG2000.
With multi-threading, FlowRoI reaches approximately 30 FPS on the same CPU.
In contrast, methods such as \cite{han2006image} require training with support vector machines, while \cite{li2023RoI} uses Swin Transformer blocks for RoI encoding, requiring large training datasets and GPU resources.
Many recent approaches rely on deep learning, achieving strong PSNR improvements but at the cost of complex training procedures and limited deployability due to GPU dependencies.
In real-world applications, ease of deployment and processing speed are equally critical.
FlowRoI satisfies these requirements but is tailored primarily for microscopy images with sparse, moving targets.
It may generalize to similar applications, but its applicability is narrower than that of deep-learning-based compression methods designed for diverse image types.

\section*{Conclusion}
We presented FlowRoI, a fast optical-flow–based region-of-interest (RoI) extraction method for high-throughput image compression in immune cell migration analysis.
By leveraging motion cues between adjacent frames, FlowRoI identifies RoIs that cover nearly all migrating cells while keeping RoI size compact.
Combined with JPEG2000 RoI compression, FlowRoI achieves 2–2.2× higher compression at equal PSNR and 2.2–2.3× higher compression at equal segmentation performance compared with standard JPEG2000.
FlowRoI runs extremely fast (30 FPS on an Intel i7-1255U laptop CPU), is training-free, requires only a few hyperparameters, and exhibits stable performance across parameter settings.
Its simplicity, robustness, and computational efficiency make FlowRoI a practical solution for reducing storage requirements in large-scale cell migration experiments while preserving the essential features required for downstream analysis.






\section*{Acknowledgements}
This research work was funded by the Deutsche Forschungsgemeinschaft (DFG), research grant GU 769/10-1 and GU769/15-1 and 15-2 (Immunostroke) and the CRC TRR332 (project C6) to M.G, as well as CH 3328/3-1 to J.C. The work of ISAS was supported by the “Ministerium für Kultur und Wissenschaft des Landes Nordrhein-Westfalen” and “Der Regierende Bürgermeister von Berlin, Senatskanzlei Wissenschaft und Forschung.”
This project was also supported by Humboldt Research Foundation.

\section*{Competing interests}
The author(s) declare no competing interest.

\section*{Ethics declarations}
NA


\section*{Data and code availability}
Code will be uploaded to GitHub once the paper has been conditionally accepted.
Correspondence and requests for materials and code could be addressed to Dr. Xiaowei Xu.



\bibliography{sample}

\end{document}